\pdfoutput=1

\documentclass[11pt]{article}

\usepackage[preprint]{acl}
\usepackage{url}
\usepackage{times}
\usepackage{latexsym}
\usepackage{amsmath} 
\usepackage{algorithm} 
\usepackage{algpseudocode} 
\usepackage[T1]{fontenc}
\usepackage{amssymb}
\usepackage[utf8]{inputenc}
\usepackage{verbatim}
\usepackage{tcolorbox}
\usepackage{listings}
\usepackage{lipsum}
\usepackage{placeins}

\usepackage{microtype}
\usepackage{hyperref}
\usepackage{url}
\usepackage{inconsolata}
\usepackage{xspace}
\usepackage{graphicx}
\usepackage{longtable}
\usepackage{array}
\usepackage{enumitem}
\usepackage{booktabs}
\usepackage{geometry}

\newcommand{\method}{\texttt{TravelAgent}\xspace}

\title{TravelAgent: An AI Assistant for Personalized Travel Planning}

\author{
Aili Chen\textsuperscript{\rm$\spadesuit$},
Xuyang Ge\textsuperscript{\rm$\spadesuit$},
Ziquan Fu\textsuperscript{\rm$\heartsuit$},
Yanghua Xiao\textsuperscript{\rm$\spadesuit$},
Jiangjie Chen\textsuperscript{\rm$\spadesuit$}\\
\textsuperscript{\rm$\spadesuit$}Fudan University\quad
\textsuperscript{\rm$\heartsuit$}System Inc.\\
\texttt{\{alchen20,xyge20,shawyh,jjchen19\}@fudan.edu.cn} \quad \texttt{frank@system.com}
}

\begin{document}
\maketitle
\begin{abstract}
As global tourism expands and artificial intelligence technology advances, intelligent travel planning services have emerged as a significant research focus. Within dynamic real-world travel scenarios with multi-dimensional constraints, services that support users in automatically creating practical and customized travel itineraries must address three key objectives: Rationality, Comprehensiveness, and Personalization. However, existing systems with rule-based combinations or LLM-based planning methods struggle to fully satisfy these criteria. 
To overcome the challenges, we introduce \method\footnote{\url{https://travel.okagent.org}}, a travel planning system powered by large language models (LLMs) designed to provide reasonable, comprehensive, and personalized travel itineraries grounded in dynamic scenarios. 
\method comprises four modules: Tool-usage, Recommendation, Planning, and Memory Module. 
We evaluate \method's performance with human and simulated users, demonstrating its overall effectiveness in three criteria and confirming the accuracy of personalized recommendations.

\end{abstract}

\section{Introduction}
\label{sec:introduction}
With the growth of global tourism, the demand for automated travel itinerary generation services has surged. Concurrently, travelers seek diverse and personalized experiences more and more. 
Early platforms~\cite{wikitravel,FrommersTravelGuides} offer static guidance without recommendation or planning. Later systems~\cite{Roadtrippers,Expedia,Booking} provide categorized recommendation and booking services but heavily rely on manual selection and rule-based combinations, unable to autonomously generate itineraries and adapt to dynamic scenarios.

Recent advancements in artificial intelligence, particularly LLMs, have promoted the development of intelligent travel services. However, despite their effectiveness in traditional tasks, LLMs face challenges in dynamic, real-world applications such as personalized travel planning, which requires considering multiple constraints and real-time information. The TravelPlanner benchmark~\cite{xie2024travelplanner} assesses language agents' capabilities in managing these constraints using a dataset of travel scenarios. Results demonstrate that even state-of-the-art LLMs~\cite{openai2022chatgpt, achiam2023gpt, TeamG2023Gemini,jiang2023mistral,jiang2024mixtral} and planning strategies~\cite{wei2022chain, yao2022react, shinn2024reflexion} struggle to meet all constraints and produce viable plans.

Despite rationality considerations, travel services designed for real-world scenarios require comprehensive itineraries incorporating dynamic updates, fine-grained details, and extensive contextual information. Additionally, personalization is also essential for enhancing the quality of services. In this study, we establish three evaluation criteria for itineraries generated by travel planning: \textbf{Rationality}, \textbf{Comprehensiveness}, and \textbf{Personalization}. Each category introduces  implementation challenges that travel service systems must address:
\begin{enumerate}[noitemsep]
     \item \textbf{Rationalily} - How to model constraints in travel scenarios and plan rational itineraries under these constraints? 
    \item \textbf{Comprehensiveness} - How to provide comprehensive travel services, including recommendation and planning, make real-time, fine-grained, and interesting itineraries?
    \item \textbf{Personalization} - How to uncover and leverage implicit user personalization information to deliver personalized recommendations and planning services?
\end{enumerate}

In this paper, we propose \method, a novel travel service system powered by LLMs and an advanced data processing framework to tackle the outlined core challenges and meet the established evaluation criteria. 
\method features four distinct modules: the \textbf{Tool-Usage Module}, \textbf{Recommendation Module}, \textbf{Planning Module}, and \textbf{Memory Module}; each is meticulously designed and collaborates seamlessly with others. 
We model the travel constraints and devise efficient algorithms to enhance the Rationality of the planning stage. 
Additionally, the Tool-usage Module encompasses a pool of real-time tools; we address the comprehension issue with history insights in the Memory Module. 
Finally, we present a new personalized recommendation framework for LLM-based agents to promote Personalization.
    
Furthermore, we rigorously evaluate the effectiveness and quality of \method through two distinct methodologies: \textbf{Overall Evaluation with human users} and \textbf{Personalization Evaluation with simulated users}. Overall Evaluation assesses \method in terms of the three criteria through detailed case studies. 
Personalization Evaluation delves deeper into Personalization, using simulated users to produce streaming behavioral data for predictive tasks in attraction recommendations. Findings indicate that \method consistently delivers high performance in Rationality, Comprehensiveness, and Personalization across multiple travel scenarios.

Our \textbf{contributions} include:
\begin{itemize}[noitemsep]
    \item\textbf{(System)} We define three objective criteria for travel services and create \method, a novel AI-driven system designed for automated personalized travel planning service to address the objectives. 
    \item \textbf{(Method)} For Rationality, we model travel constraints and develop a spatiotemporal-aware planning algorithm. For Comprehensiveness, we establish a real-time data mechanism enhanced by multiple tools. For Personalization, we introduce an agent-based personalized recommendation framework with continuous learning of implicit information.
    \item \textbf{(Evaluation)} We conduct human case studies to evaluate the overall performance of \method, which surpasses the baseline GPT-4+ agent across three criteria. Further, we undertake behavioral prediction tasks with simulated users and find that our personalized recommendation framework exhibits lower error rates than baseline LLM-based recommendation methods.

\end{itemize}

\section{Related Work}
\label{sec:related}

\paragraph{LLM-based Agent} LLM-based agents have demonstrated significant capability  in executing multi-step tasks~\cite{wang2024survey}, showcasing their potential in fields such as education~\cite{wang2024survey, swan2023math}, healthcare~\cite{zhang2023huatuogpt}, and emotional support~\cite{hasan2023sapien}. To enhance their planning ability, some researchers propose task decomposition methods~\cite{shen2024hugginggpt,wang2023plan,singh2023progprompt}, while others focus on step-by-step reasoning policies for dynamic planning~\cite{wei2022chain,yao2024tree,besta2024graph,chen2022program,yao2022react}. Additionally, through memory construction~\cite{zhong2024memorybank, liu2023think,  hu2023chatdb, zheng2023synapse, huang2023memory} and tool use~\cite{schick2024toolformer, qin2023toolllm, lu2024chameleon, wang2024llms, yuan2024easytool}, LLM-based agents can leverage historical and external resources to better adaptation, making it a promising tool for travel services.
\paragraph{LLM as Recommendation System} With the advent of LLMs, researchers endeavor to leverage LLMs for direct recommendations. Two main approaches have emerged: Direct Recommendation~\cite{dai2023uncovering, hou2024large, kang2023llms, wang2023zero} and Decode and Encode~\cite{zhou2024language, yang2023palr, liu2023chatgpt}. The former prompts LLMs with users' behavioral histories to recommend directly, while the latter decodes a one-time user profile from the last interaction for subsequent recommendations. However, both approaches have limitations in capturing long-term user preferences through multiple interactions, resulting in suboptimal effectiveness.

\section{The TravelAgent System Design}
\label{sec:method}
\method system is architected into five key modules: (1) \textbf{Tool-usage Module}, which integrates various utilities drawing from real-time dynamic data sources or arithmetic algorithms; (2) \textbf{Recommendation Module}, which delivers personalized travel recommendations tailored to individual preferences; (3) \textbf{Planning Module}, which incorporates budget and route planning functionalities, leveraging budget plan result to guide detailed route plans sensitive to time and geographic constraints; (4) \textbf{Memory Module}, which help to continually learn user preferences as a comprehensive repository of travel knowledge and user profiles.

\subsection{Constraints Modeling}
\begin{figure}
    \centering
    \includegraphics[width=0.9\linewidth]{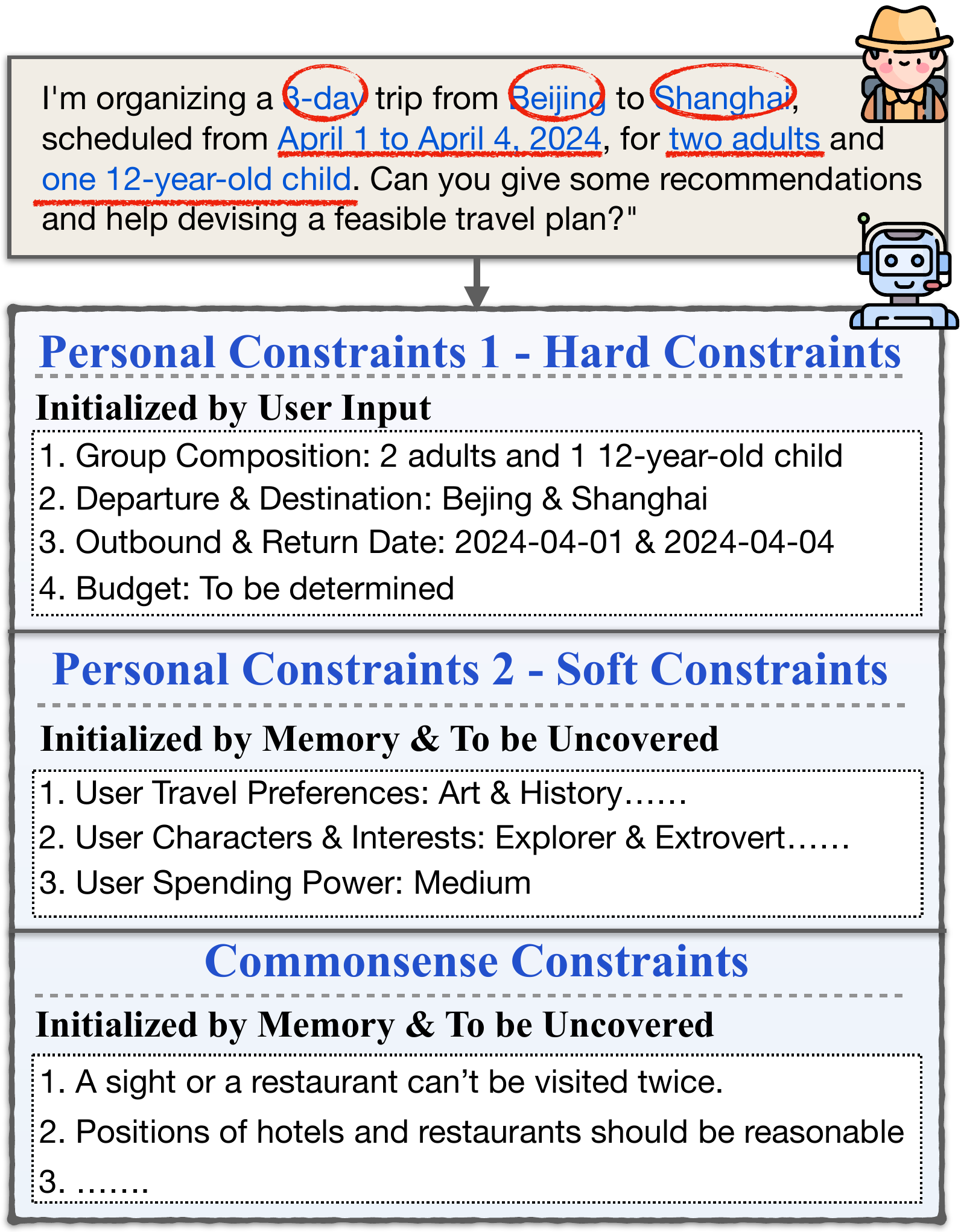}
    \caption{Initial constraints modeling with user input and Memory Module.}
    \label{fig:initial_constraints_modeling}
\end{figure}

The core challenge in travel services lies in handling multi-dimensional constraints in complex real-world travel scenarios. In this work, we categorize these constraints into - \textbf{(1) Personal Constraints}, which are unique to individual users and include Hard Constraints and Soft Constraints; \textbf{(2) Commonsense Constraints}, which encompass general travel knowledge applicable across various scenarios and beneficial to all users. Figure~\ref{fig:initial_constraints_modeling} illustrates an initial constraints example.
\begin{itemize}[noitemsep]
     \item \textbf{Personal Constraints 1 - Hard Constraints} are both user and scenario-specific, which must be strictly followed. These non-negotiable elements, such as the outbound/return date, are directly provided by users in the current scenario and reflect their travel objectives, which are essential for Rationality and Personalization of \method.
    \item \textbf{Personal Constraints 2 - Soft Constraints} are user-specific but cross-scenarios, initiated by historical insights from Memory Module. Instead of explicitly provided by users, they are uncovered gradually by \method across multiple scenarios and evolve with user interactions. These personalized elements are not mandatory but significantly affect Personalization and user satisfaction.
    \item \textbf{Commonsense Constraints} are general knowledge applicable to users and scenarios, which are initiated by Memory Module and continually learned by \method. They are crucial for Rationality of itineraries and travel experiences, such as preventing the repetition of dining restaurants.
\end{itemize}

\begin{figure}
    \centering
    \includegraphics[width=0.9\linewidth]{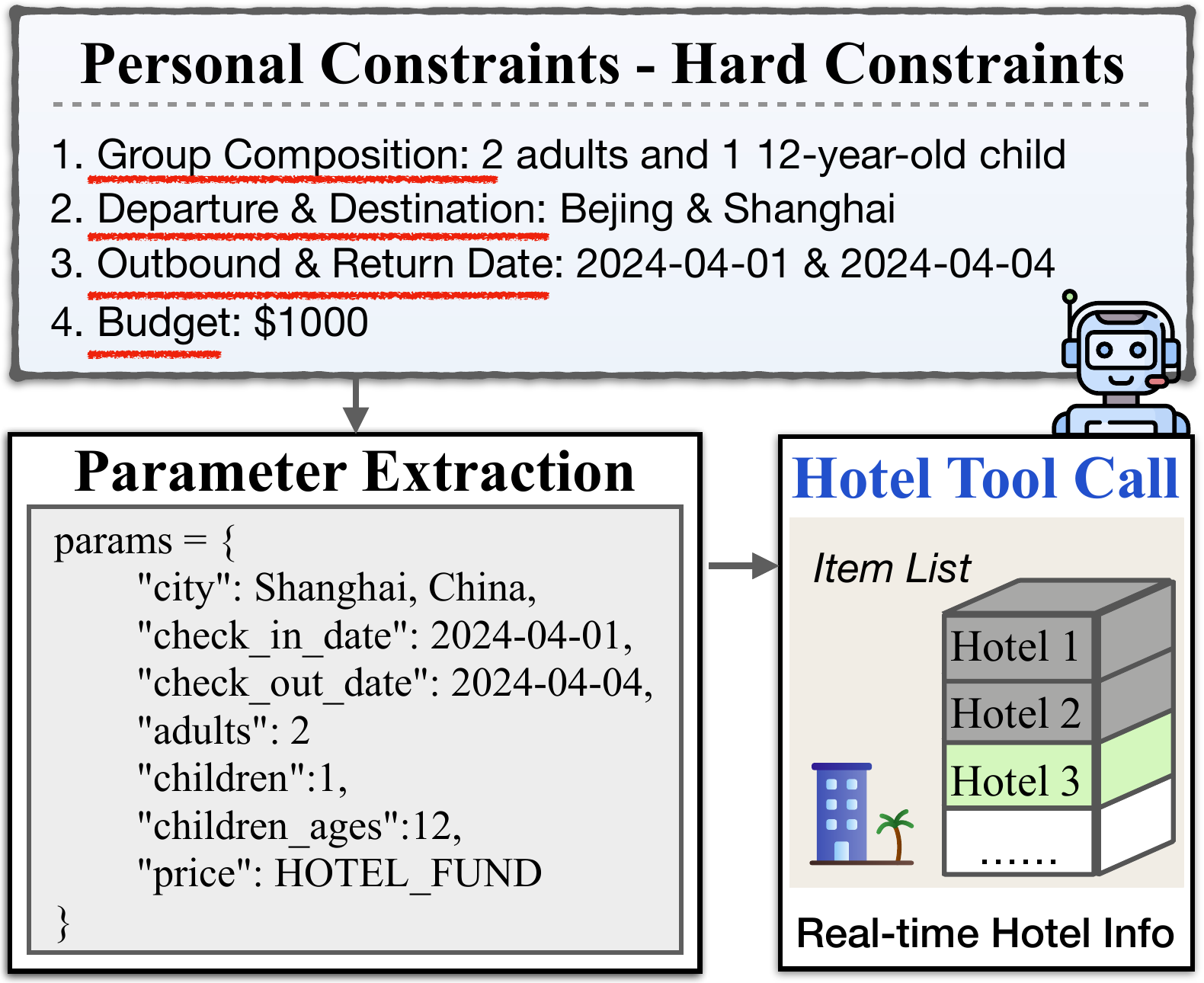}
    \caption{An example of Hotel Tool call. Appendix \ref{appendix:tools} provides a comprehensive overview of all types and functions of tools.}
    \label{fig: An example of hotel tool call}
\end{figure}

\subsection{Tool-usage Module}
\method integrates various real-time data sources to ensure up-to-date recommendations and plans. 
Tool-usage Module, which underpins Rationality and Personalization and serves as the core component for enhancing Comprehensiveness, is divided into two categories: \textbf{real-time API tools} and \textbf{arithmetic algorithm tools}. The first category, powered by SerpAPI\footnote{\url{https://serpapi.com/integrations/python\#google-search-api-capability}} and Google Maps API\footnote{\url{https://developers.google.com/maps}}, includes specialized utilities such as Attraction Tool, designed to fetch high-quality real-time data. The second category comprises tools like Distance and Time Tools, which employ arithmetic algorithms to ensure precise route planning.

As shown in Figure~\ref{fig: An example of hotel tool call}, after initial constraints modeling, Tool-usage Module extracts structured parameters from Hard Constraints, serving as the principal filter at the tool-call stage. Subsequently, it invokes specific tools according to the targeted recommendation or planning point. The filtered information is then transmitted to Recommendation Module for additional refinement and finally, conveyed to Planning Module.

\subsection{Recommendation Module}
Recommendation Module serves as the primary component to achieve Personalization of \method. It diverges from traditional paradigms of recommendation systems by directly leveraging LLM-based agents to perform online recommendations. Inspired by some related work for online Personalization construction~\cite{chidlovskii2015improved, dalvi2022towards, baek2024knowledge, ma2024beyond}, we form a data-driven recommendation framework comprises Budget, City, Flight, Hotel, Restaurant, and Attraction Recommender, which continually learns implicit personalization information through multiple interactions. 

Recommendation Module capitalizes on four distinct data sources for recommendations: 
\begin{itemize}[noitemsep]
    \item \textbf{In-context Travel Constraints}, sourced from the user interface, refers to Hard Constraints provided by users, reflecting their specific needs for the current travel scenario. 
    \item \textbf{Real-time Travel Information}, derived from \textbf{Tool-usage Module}, encompasses up-to-date details on various travel options, each characterized by multi-dimensional attributes.
    \item \textbf{History Insights}, retrieved from \textbf{Memory Module}, comprises learned commonsense knowledge, user personas, and past interactions records. These insights dynamically evolve through new interactions and finally re-storing in Memory Module for future use.
    \item \textbf{Generalized Knowledge}, sourced from \textbf{LLMs}, incorporates a wide range of fundamental travel-related knowledge.
\end{itemize}

\begin{figure}
    \centering
    \includegraphics[width=\linewidth]{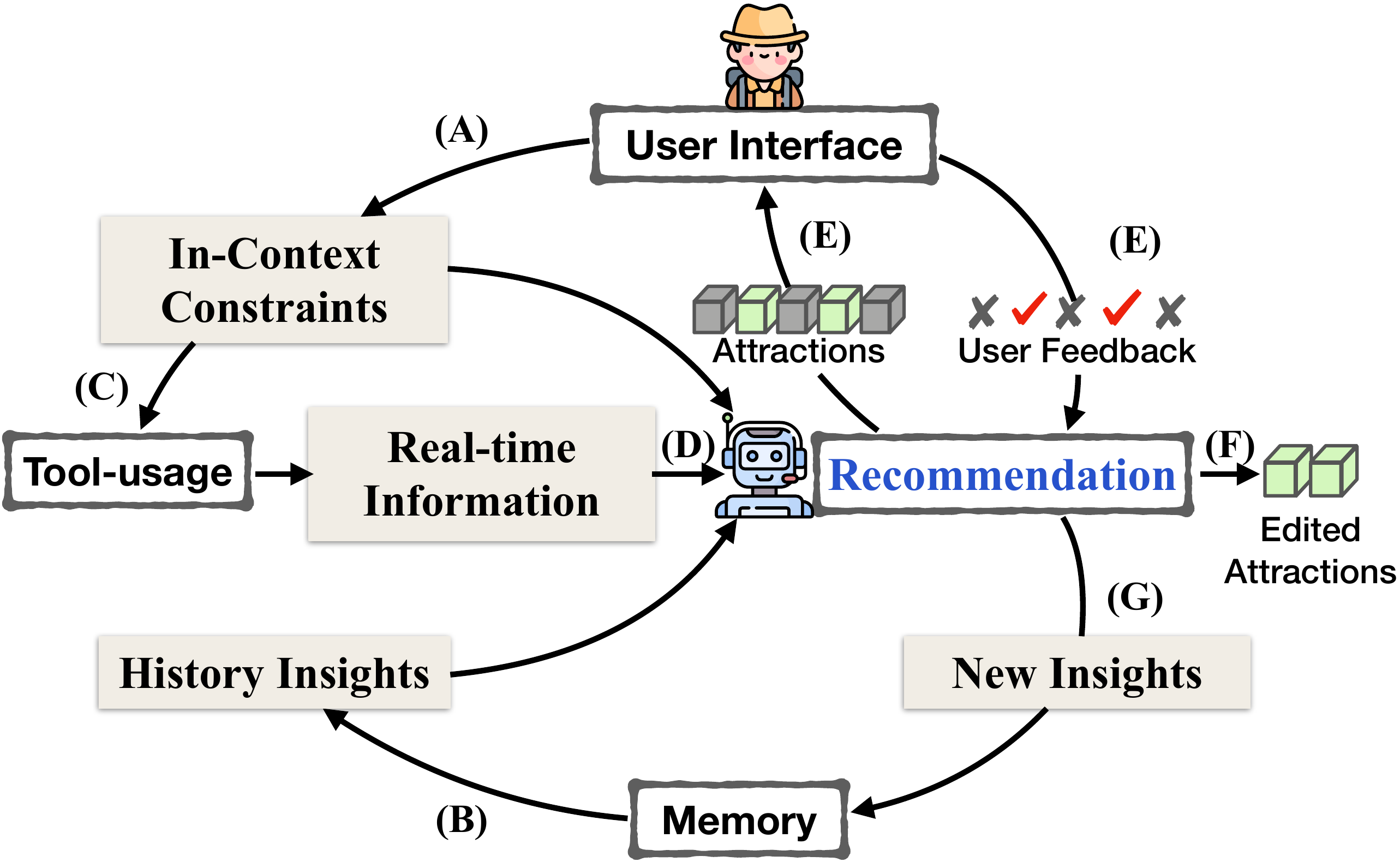}
    \caption{The process of Attraction Recommendation.}
    \label{fig:attraction_recommendation}
\end{figure}

Take attraction recommendation as an instance, the whole worflow is shown in Figure \ref{fig:attraction_recommendation} - \textbf{(A) In-context Constraints Extraction}: At the start of a travel scenario, Hard Constraints are initialized based on the user's input from the user interface; \textbf{(B) History Insights Retrieval}: Concurrently, the system retrieves relevant historical insights from Memory Module, initializing implicit constraints, including both soft and commonsense constraints; \textbf{(C) Real-time Information Gathering}: At the attraction recommendation stage, Attraction Tool uses Hard Constraints to fetch up-to-date attraction information; \textbf{(D) Data-driven Recommendation}: Recommendation Module synthesizes all data and leverage LLMs to generate personalized attraction recommendations and visit suggestions; \textbf{(E) User Behavioral Interactions}: Recommendation list, accompanied by contextual information, are presented on the structured user interface for user feedback (like/pass); \textbf{(F) Items Re-rank} Based on user feedback, the Recommendation Module re-ranks the attractions, finalizing the selection for route planning; \textbf{(G) Implicit Insight Learning}: Post-interaction, the system draws new short-term insights relevant to users' persona and commonsense knowledge, updating long-term insights in Memory Module. With this pipeline, \method can continually learn implicit knowledge and dynamically retain a tailored insight of each user.

\begin{figure*}[h]
    \centering
    \includegraphics[width=1\linewidth]{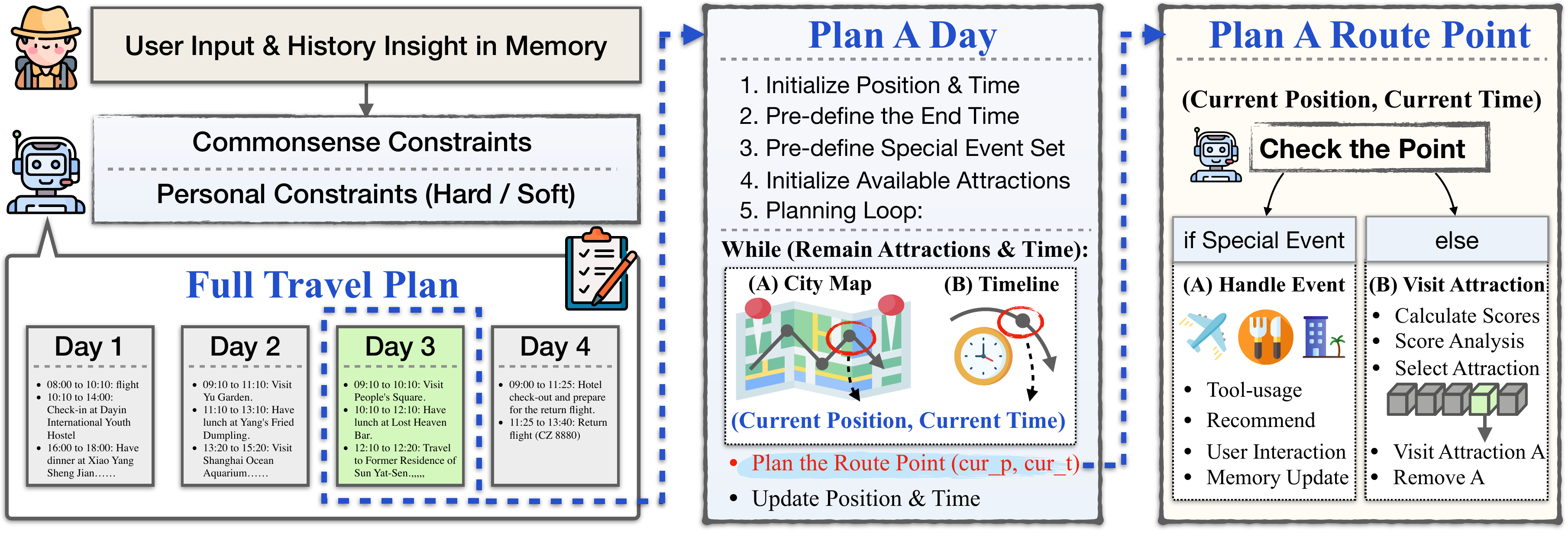}
    \caption{The whole workflow of Route Planner illustrates the dynamic generation process of multi-day itineraries.}
    \label{fig:route_plan_workflow}
\end{figure*}

\subsection{Planning Module}
Planning Module comprises \textbf{Budget Planner} and \textbf{Route Planner}. We innovatively integrate budgetary elements into the planning process to enhance the itinerary's Rationality and Comprehensiveness, empowering planning with LLMs and a spatiotemporal-aware route algorithm.
\paragraph{Budget Planner}
After initial constraint modeling, Budget Recommender engages LLMs to generate a preliminary budget estimate. Once users confirm their budget, Budget Planner re-evaluates users' financial situation and strategically distributes the budget across six categories: Accommodation, Attractions, Restaurants, Transportation, Other Expenses, and a Reserve Fund, primarily adheres to commonsense knowledge and dynamically adjusts according to other constraints to ensure that the travel plans are economically viable and tailored to the context.

\paragraph{Route Planner}
As shown in Figure \ref{fig:route_plan_workflow}, guided by the budget plan, multi-day itineraries are dynamically generated using an attraction set provided by Recommendation Module. Each day's planning starts by setting initial conditions, including current position $p_{cur}$, current time $t_{cur}$, and an end time. In addition, we pre-defined some anticipated event routines according to Commonsense Constraints to ensure rational planning, such as dining or check-ins. 
The daily planning process iterates as long as time permits and attractions are available, with each iteration re-evaluating the route point characterized by the current coordinates $(p_{cur}, t_{cur})$. If the current coordinates align with a special event, corresponding actions are executed, such as selecting an optimal nearby restaurant. Otherwise, the planner evaluates the available attractions and selects one for a visit. After each route point, the coordinates $(p_{cur}, t_{cur})$ are updated to align subsequent activities with the strategic temporal and spatial logistics.

\paragraph{Attraction Scores Calculation}
For better assessment at $(p_{cur},t_{cur})$, we devise Algorithm \ref{alg:calc_score_attraction} to score available attractions. For each attraction $A_i$, firstly check the distance $d_i$ from $p_{cur}$ to the attraction $p_i$ to compute three time-(1) the travel time $t_{cost,i}$, including both the travel duration at speed $v$ and the recommended visit duration $\delta t_{\text{rec},i}$; (2) the arrival time $T_{arr,i}$ (3) the leave time $T_{dep,i}$. The scoring mechanism involves three components:
\begin{itemize}[noitemsep]
\item \textbf{Return Time Score} $S_{\text{ret}, i}$, which checks if the total travel time and duration at the $A_i$ fit within the remaining day time $t_{\text{left}}$.
\item \textbf{Optimal Visit Time Score} $S_{\text{opt}, i}$, which evaluates how well the visit aligns with the preferred visiting window $\tau_i$.
\item \textbf{Reserve Time Score} $S_{\text{left}, i}$, which optimizes the use of the remaining time, prioritizing efficient day planning.
\end{itemize}

The total score $S_i = S_{\text{ret},i} + S_{\text{opt},i} + S_{\text{left},i}$ guides the selection of the next attraction, ensuring the itinerary is balanced, enjoyable, and logistically sound. This method dynamically adapts to daily circumstances, aligning travel plans with real-time constraints and preferences.

\begin{algorithm}[t]
\caption{Calculate Score for Attraction $A_i$}
\label{alg:calc_score_attraction}
\begin{algorithmic}[1]
\footnotesize
\Require 
\Statex \textbf{Current Position} $p_{cur}$, \textbf{Current Time} $t_{cur}$
\Statex \textbf{Available Attraction} $A_i = (p_i, \delta t_i, \tau_i)$
\Statex \Comment{\text{Attraction Position} $A_i$}
\Statex \Comment{\text{Recommended Duration} $\delta t_i$}
\Statex \Comment{\text{Recommended Visit Time Window $\tau_i$}}
\Statex \textbf{Left Time} $t_{left}$, \textbf{Travel Speed} $v$

\Ensure $S_i$
\State $d_i \gets \|p_{cur} - p_i\|$ \Comment{\text{Calculate distance}}
\State $t_{cost,i} \gets \frac{d_i}{v} + \delta t_{rec,i}$ \Comment{\text{Calculate the travel time}}
\State $T_{arr,i} \gets t_{cur} + \frac{d_i}{v}$ \Comment{\text{Calculate the arrival time}}
\State $T_{dep,i} \gets T_{arr,i} + \delta t_i$ \Comment{\text{Calculate the leave time}}
\State $S_{ret,i} \gets \Call{ReturnTimeScore}{t_{left}, t_{cost,i}}$
\State $S_{opt,i} \gets \Call{VisitTimeScore}{(T_{arr,i}, T_{dep,i}), \tau_i}$
\State $S_{left,i} \gets \Call{ReserveTimeScore}{t_{left}, t_{cost,i}}$
\State $S_i \gets S_{ret,i} + S_{opt,i} + S_{left,i}$ \Comment{\text{Final score}}
\State \textbf{return} $S_i$
\end{algorithmic}
\end{algorithm}

\begin{figure*}[t]
    \centering
    \includegraphics[width=1\linewidth]{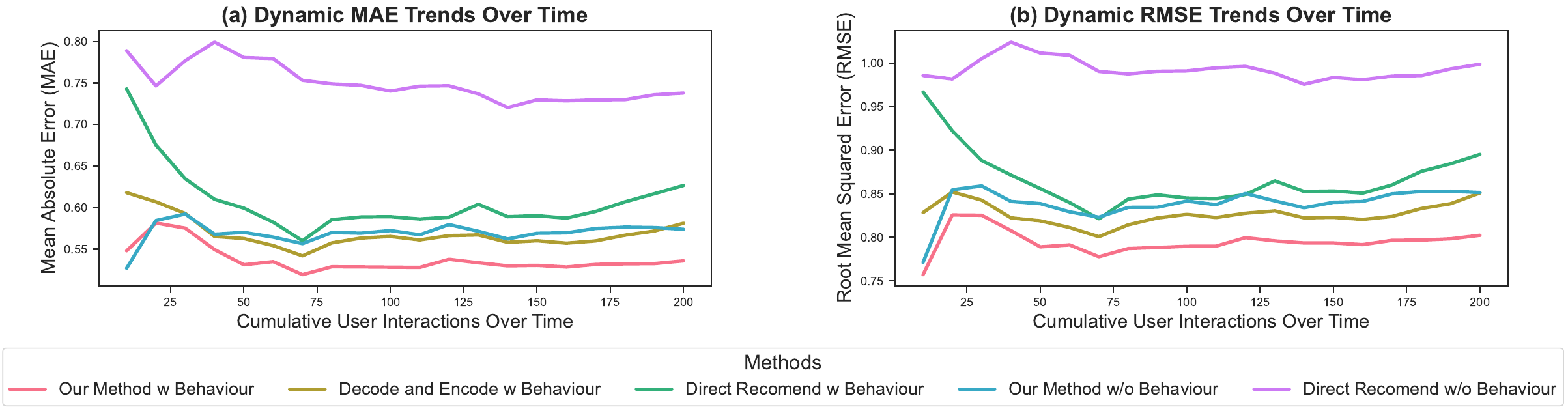}
    \caption{MAE and RMSE trends over simulated user interactions. When with user behavior history, our methods demonstrate superior cold-start performance and excellent error control. When without user behavior history, our methods mitigate performance degradation effectively compared to baselines. Prompts are shown in Appendix \ref{appendix:prompts}.}
    \label{fig:MAE_RMSE_trends}
\end{figure*}

\subsection{Memory Module}
Memory Module is designed to promote Personalization and dynamic data processing and consists of two primary components: (1) \textbf{Short-term Memory} stores new user interactions and captures the immediate contextual travel data, constraints, and plans, adapting to the latest user preferences. (2) \textbf{Long-term Memory} retains user data over extended periods, accumulating a comprehensive persona based on historical interactions. It stores stable user preferences, such as a user's consistent interest in eco-friendly attractions or specific dietary requirements. Historical insights allow the system to provide recommendations that align with long-term travel habits and preferences. Examples of long-term insights are in Appendix \ref{appendix:insights}.

\section{Evaluation}
\label{sec:evaluation}

We conduct simulated and human user evaluations of \method to verify its performance against the three criteria: Rationality, Comprehensiveness, and Personalization.

\subsection{Overall Evaluation}

\begin{table}[t]
  \centering
  \small
  \begin{tabular}{p{2.2cm}cc}
    \toprule
    \textbf{Criteria} & \textbf{\method} & \textbf{GPT-4+ agent}\\
    \midrule
    Rationality & 9.56 &8.16\\
    Comprehensiveness & 8.87 &  6.25\\
    Personalization &8.44  &4.31\\
    \bottomrule
  \end{tabular}
  \caption{\label{tab:human_evaluation_result} Average human evaluation scores of Rationality, Comprehensiveness, and Personalization for \method and GPT-4+ agent across 20 travel cases.}
\end{table}

This section presents the results from case studies that evaluated \method using 20 distinct travel scenarios and user input forms (shown in Appendix \ref{appendix:forms}). To establish a baseline, these case studies were also conducted with a typical agent system, the GPT-4+ agent~\cite{achiam2023gpt}, to showcase the performance of the proposed \method. In this section, human users interact with \method and the GPT-4+ agent online using these 20 input forms. Throughout the entire workflow and analysis of generated itineraries, human users evaluated three criteria on a scale from 1 to 10: Rationality, Comprehensiveness, and Personalization. Each criterion was assessed according to a predefined set of evaluation dimensions detailed in Appendix \ref{appendix:evals}. The average scores are presented in Table~\ref{tab:human_evaluation_result}, and comparative case study examples are detailed in Appendices \ref{appendix:case1} and \ref{appendix:case2}.

\subsection{Personalization Evaluation}
To assess Personalization Module of \method and the performance of our agent-based online recommendation framework, we experiment with 100 simulated users based on GPT-3.5-turbo-0125 ~\cite{openai2022chatgpt} and 10 travel scenarios and instruct Recommendation Module to suggest 20 attractions per city to each user. Simulated users rate items based on preferences, and \method predicts these ratings while continually learning over interactions. We compared our method with two popular LLM-based recommendation frameworks: (1) \textbf{Direct Recommend} and (2) \textbf{Decode and Encode}, both lacking memory and continual learning. Evaluation metrics included Mean Absolute Error (MAE) and Root Mean Squared Error (RMSE). Figure \ref{fig:MAE_RMSE_trends} shows the trend results over user interactions and our methods' high prediction accuracy, verifying \method's Personalization.

\section{Conclusion}
\label{sec:conclusion}
In this work, we introduced \method, a novel travel service system leveraging LLMs and real-time data for automated itinerary generation in real-world scenarios with multi-dimensional constraints. We establish three criteria for travel planning and model travel constraints. To address the challenges, we developed tailored spatiotemporal-aware algorithms and a novel agent-based personalized online recommendation framework enhanced by a real-time data processing mechanism and dynamic memory. We assess \method through simulated user rating prediction and real-world human case studies. The system delivers rational, comprehensive, and personalized travel services. We hope our work will inspire further exploration of travel services and LLM-based agent applications.

\section*{Limitations}
\label{sec:limitation}
While \method we proposed aims to address rational and personalized travel itinerary generation for real-world travel scenarios, it is essential to acknowledge certain limitations in our study. Firstly, the system heavily relies on the quality of dynamic real-time data sources; inaccuracies, such as outdated flight information, can result in suboptimal recommendations and impractical final plans. Future research should focus on enhancing the quality and reliability of these data sources to improve adaptability across different travel scenarios. Secondly, a deeper exploration of personalized activities in travel itineraries based on more detailed user personalization information is needed for further improvement.

\bibliography{custom}

\clearpage
\appendix

\begin{appendix}
\label{sec:appendix}
\section{Tool-usage Module Details}
\label{appendix:tools}

\begin{table}[h]
  \centering
  \small
  \begin{tabular}{llll}
    \toprule
    \textbf{Tool} & \textbf{Function} & \textbf{Type} & \textbf{Description} \\
    \midrule
    City Tool & get\_city\_info & API-based &Retrieve real-time city information \\
    Flight Tool & get\_flight\_info & API-based &Retrieve real-time flight information \\
    Hotel Tool & get\_hotel\_info & API-based &Retrieve real-time hotel information \\
    Restaurant Tool & get\_restaurant\_info & API-based &Retrieve real-time restaurant information \\
    Sight Tool & get\_sight\_info & API-based &Retrieve real-time attraction information \\
    Tip Tool & get\_tip\_info & API-based & Retrieve travel tips and guides \\
    Direction Tool & get\_direction\_info & API-based &Retrieve route traffic information \\
    Position Tool & get\_position\_info & API-based &Retrieve target position information \\
    Distance Tool & get\_distance\_info & Algorithm-based &Calculate distance \\
    Time Tool & get\_time\_info & Algorithm-based &Time conversion\\
    \bottomrule
  \end{tabular}
  \caption{\label{tab:tool_modules} List of Tools in Personalized \method}
\end{table}

\section{Long-term History Insight Examples}
\label{appendix:insights}
\begin{itemize}
\item\textbf{Soft Constraints Insights}

\begin{tabular}{|p{14cm}|}
\hline
\begin{lstlisting}[basicstyle=\scriptsize, breaklines=true, belowskip=0pt]
"{"user_id": 2; "user_spending_level": "low", "user_attraction_preference": "The user has shown a preference for free or low-cost attractions that are suitable for both adults and a child. They have liked sights that offer cultural experiences, historical insights, and opportunities for leisure and relaxation in urban green spaces. The user has shown a preference for affordable or free attractions that are suitable for families and can be enjoyed by a young child. They have liked sights that offer a mix of cultural, historical, and recreational experiences, such as classical gardens, bustling squares, and pedestrian streets with vibrant atmospheres.", "user_restaurant_preference": "Chinese"......}"
\end{lstlisting} \\
\hline
\end{tabular}

\item\textbf{Commonsense Constraints Insights}
\begin{tabular}{|p{14cm}|}
\hline
\begin{lstlisting}[basicstyle=\scriptsize, breaklines=true, belowskip=0pt]
"If low spending level and family tour: Considering the user's low spending level and the presence of a young child, attractions should be family-friendly, free or low-cost, and allow for short visits that accommodate a child's attention span. The weather should be taken into consideration, outdoor activities should be planned for cooler parts of the day, and indoor or shaded areas should be prioritized during peak heat hours. Attractions should also be geographically practical, minimizing travel time and maximizing sightseeing time."
\end{lstlisting} \\
\hline
\end{tabular}

\end{itemize}

\clearpage

\section{Baseline Recommendation Prompts}
\label{appendix:prompts}
\begin{tcolorbox}[fonttitle=\bfseries, boxrule=0.5mm, width=\textwidth, sharp corners=south]
\lstset{
    basicstyle=\ttfamily\footnotesize,
    breaklines=true,
    columns=fullflexible
}
\begin{lstlisting}
DIRECT_PREDICT_TEMPLATE = """
    Task: rating prediction
    The user has rated the {item_type}:
    {previous_user_behaviours}
    Based on this user rating history, guess what score out of 5 would the user give to the following {item_type}:
    {recommend_item_list}.
    The number of your choices MUST be the same as the number of candidate {item_type} and your choices order should be consistent with the given order of {item_type}.
""".strip()
\end{lstlisting}
\end{tcolorbox}

\begin{tcolorbox}[fonttitle=\bfseries, boxrule=0.5mm, width=\textwidth, sharp corners=south]
\lstset{
    basicstyle=\ttfamily\footnotesize,
    breaklines=true,
    columns=fullflexible
}
\begin{lstlisting}
PROFILE_DECODE_TEMPLATE = """
    You have recommended some {item_type} to the user:
    {recommend_items_list}
    I have rated the {item_type}:
    {user_behaviours}
    Based on this rating history, summarize the reasons why I like or dislike certain {item_type} in under 100 words. Do not quote {item_type} titles.
""".strip()
\end{lstlisting}
\end{tcolorbox}

\begin{tcolorbox}[fonttitle=\bfseries, boxrule=0.5mm, width=\textwidth, sharp corners=south]
\lstset{
    basicstyle=\ttfamily\footnotesize,
    breaklines=true,
    columns=fullflexible
}
\begin{lstlisting}
PROFILE_ENCODE_WITH_HISTORY_TEMPLATE = """
    This is your profile: {user_profile}.
    You has rated the {item_type}:
    {previous_user_behaviours}
    Based on this, what score out of 5 would you give to the following {item_type}:
    {recommend_item_list}.
""".strip()
\end{lstlisting}
\end{tcolorbox}

\begin{tcolorbox}[fonttitle=\bfseries, boxrule=0.5mm, width=\textwidth, sharp corners=south]
\lstset{
    basicstyle=\ttfamily\footnotesize,
    breaklines=true,
    columns=fullflexible
}
\begin{lstlisting}
PROFILE_ENCODE_WITHOUT_HISTORY_TEMPLATE = """
    This is your profile: {user_profile}.
    What score out of 5 would you give to the following {item_type}:
    {recommend_item_list}.
""".strip()
\end{lstlisting}
\end{tcolorbox}

\clearpage

\section{Input Form Cases}
\label{appendix:forms}

\begin{tabular}{|>{\scriptsize}m{16cm}|}
\hline
\begin{lstlisting}[basicstyle=\scriptsize, breaklines=true,  belowskip=0pt]
{"user_name": "sheep", "departure": "beijing, beijing, china", "destination": "shanghai, shanghai, china", "outbound_date": "2024-08-06", "return_date": "2024-08-09", "adults_num": 2, "children_num": 1, "children_ages": "3", "restaurant_preference": "Chinese"}
\end{lstlisting} \\
\begin{lstlisting}[basicstyle=\scriptsize, breaklines=true,   aboveskip=0pt, belowskip=0pt]
{"user_name": "Amy", "departure": "new york, new york, usa", "destination": "los angeles, california, usa", "outbound_date": "2024-09-01", "return_date": "2024-09-03", "adults_num": 1, "children_num": 1, "children_ages": "5", "restaurant_preference": "Italian"}
\end{lstlisting} \\
\begin{lstlisting}[basicstyle=\scriptsize, breaklines=true,   aboveskip=0pt, belowskip=0pt]
{"user_name": "Bob", "departure": "paris, france, france", "destination": "lyon, auvergne-rhone-alpes, france", "outbound_date": "2024-08-17", "return_date": "2024-08-20", "adults_num": 3, "children_num": 0, "children_ages": "", "restaurant_preference": "French"}
\end{lstlisting} \\
\begin{lstlisting}[basicstyle=\scriptsize, breaklines=true,   aboveskip=0pt, belowskip=0pt]
{"user_name": "john", "departure": "london, england, uk", "destination": "edinburgh, scotland, uk", "outbound_date": "2024-09-16", "return_date": "2024-09-20", "adults_num": 2, "children_num": 1, "children_ages": "8", "restaurant_preference": "British"}
\end{lstlisting} \\
\begin{lstlisting}[basicstyle=\scriptsize, breaklines=true,   aboveskip=0pt, belowskip=0pt]
{"user_name": "lucy", "departure": "sydney, new south wales, australia", "destination": "melbourne, victoria, australia", "outbound_date": "2024-10-05", "return_date": "2024-10-06", "adults_num": 1, "children_num": 2, "children_ages": "7, 9", "restaurant_preference": "Seafood"}
\end{lstlisting} \\
\begin{lstlisting}[basicstyle=\scriptsize, breaklines=true,   aboveskip=0pt, belowskip=0pt]
{"user_name": "mike", "departure": "toronto, ontario, canada", "destination": "vancouver, british columbia, canada", "outbound_date": "2024-08-15", "return_date": "2024-08-18", "adults_num": 2, "children_num": 0, "children_ages": "", "restaurant_preference": "Steak"}
\end{lstlisting} \\
\begin{lstlisting}[basicstyle=\scriptsize, breaklines=true,   aboveskip=0pt, belowskip=0pt]
{"user_name": "anna", "departure": "berlin, berlin, germany", "destination": "munich, bavaria, germany", "outbound_date": "2024-9-09", "return_date": "2024-9-10", "adults_num": 2, "children_num": 2, "children_ages": "4, 6", "restaurant_preference": "German"}
\end{lstlisting} \\
\begin{lstlisting}[basicstyle=\scriptsize, breaklines=true,   aboveskip=0pt, belowskip=0pt]
{"user_name": "sara", "departure": "madrid, community of madrid, spain", "destination": "barcelona, catalonia, spain", "outbound_date": "2024-09-21", "return_date": "2024-09-25", "adults_num": 3, "children_num": 1, "children_ages": "10", "restaurant_preference": "French"}
\end{lstlisting} \\
\begin{lstlisting}[basicstyle=\scriptsize, breaklines=true,   aboveskip=0pt, belowskip=0pt]
{"user_name": "jose", "departure": "rio de janeiro, rio de janeiro, brazil", "destination": "sao paulo, sao paulo, brazil", "outbound_date": "2024-9-01", "return_date": "2024-9-04", "adults_num": 1, "children_num": 0, "children_ages": "", "restaurant_preference": "Chinese"}
\end{lstlisting} \\
\begin{lstlisting}[basicstyle=\scriptsize, breaklines=true,   aboveskip=0pt, belowskip=0pt]
{"user_name": "emma", "departure": "rome, lazio, italy", "destination": "florence, tuscany, italy", "outbound_date": "2024-10-02", "return_date": "2024-10-05", "adults_num": 2, "children_num": 1, "children_ages": "5", "restaurant_preference": "Italian"}
\end{lstlisting} \\
\begin{lstlisting}[basicstyle=\scriptsize, breaklines=true,   aboveskip=0pt, belowskip=0pt]
{"user_name": "li", "departure": "beijing, beijing, china", "destination": "guangzhou, guangdong, china", "outbound_date": "2024-09-05", "return_date": "2024-09-07", "adults_num": 2, "children_num": 2, "children_ages": "6, 9", "restaurant_preference": "French"}
\end{lstlisting} \\
\begin{lstlisting}[basicstyle=\scriptsize, breaklines=true,   aboveskip=0pt, belowskip=0pt]
{"user_name": "ali", "departure": "dubai, dubai, uae", "destination": "abu dhabi, abu dhabi, uae", "outbound_date": "2024-8-10", "return_date": "2024-8-11", "adults_num": 2, "children_num": 1, "children_ages": "7", "restaurant_preference": "Seafood"}
\end{lstlisting} \\
\begin{lstlisting}[basicstyle=\scriptsize, breaklines=true,   aboveskip=0pt, belowskip=0pt]
{"user_name": "hiro", "departure": "tokyo, tokyo, japan", "destination": "kyoto, kyoto, japan", "outbound_date": "2024-08-25", "return_date": "2024-08-27", "adults_num": 1, "children_num": 1, "children_ages": "3", "restaurant_preference": "Japanese"}
\end{lstlisting} \\
\begin{lstlisting}[basicstyle=\scriptsize, breaklines=true,   aboveskip=0pt, belowskip=0pt]
{"user_name": "oliver", "departure": "dallas, texas, usa", "destination": "austin, texas, usa", "outbound_date": "2024-09-01", "return_date": "2024-09-05", "adults_num": 2, "children_num": 0, "children_ages": "" "restaurant_preference": "Italian"}
\end{lstlisting} \\
\begin{lstlisting}[basicstyle=\scriptsize, breaklines=true,   aboveskip=0pt, belowskip=0pt]
{"user_name": "linda", "departure": "helsinki, uusimaa, finland", "destination": "tampere, pirkanmaa, finland", "outbound_date": "2024-08-16", "return_date": "2024-08-20", "adults_num": 1, "children_num": 1, "children_ages": "6", "restaurant_preference": "French"}
\end{lstlisting} \\
\begin{lstlisting}[basicstyle=\scriptsize, breaklines=true,   aboveskip=0pt, belowskip=0pt]
{"user_name": "nina", "departure": "moscow, moscow, russia", "destination": "st. petersburg, leningrad, russia", "outbound_date": "2024-10-12", "return_date": "2024-10-13", "adults_num": 1, "children_num": 1, "children_ages": "8", "restaurant_preference": "Russian"}
\end{lstlisting} \\
\begin{lstlisting}[basicstyle=\scriptsize, breaklines=true,   aboveskip=0pt, belowskip=0pt]
{"user_name": "peter", "departure": "amsterdam, north holland, netherlands", "destination": "rotterdam, south holland, netherlands", "outbound_date": "2024-09-03", "return_date": "2024-09-07", "adults_num": 2, "children_num": 0, "children_ages": "", "restaurant_preference": "Dutch"}
\end{lstlisting} \\
\begin{lstlisting}[basicstyle=\scriptsize, breaklines=true,   aboveskip=0pt, belowskip=0pt]
{"user_name": "lucas", "departure": "sao paulo, sao paulo, brazil", "destination": "rio de janeiro, rio de janeiro, brazil", "outbound_date": "2024-10-12", "return_date": "2024-10-15", "adults_num": 3, "children_num": 1, "children_ages": "6", "restaurant_preference": "Brazilian"}
\end{lstlisting} \\
\begin{lstlisting}[basicstyle=\scriptsize, breaklines=true,   aboveskip=0pt, belowskip=0pt]
{"user_name": "mia", "departure": "bangkok, bangkok, thailand", "destination": "chiang mai, chiang mai, thailand", "outbound_date": "2024-08-20", "return_date": "2024-08-22", "adults_num": 1, "children_num": 1, "children_ages": "4", "restaurant_preference": "Thai"}
\end{lstlisting} \\
\begin{lstlisting}[basicstyle=\scriptsize, breaklines=true,   aboveskip=0pt, belowskip=0pt]
{"user_name": "eva", "departure": "vienna, vienna, austria", "destination": "salzburg, salzburg, austria", "outbound_date": "2024-09-25", "return_date": "2024-09-28", "adults_num": 2, "children_num": 2, "children_ages": "3, 5", "restaurant_preference": "Austrian"}
\end{lstlisting} \\
\hline

\end{tabular}

\clearpage
\section{Human Evaluation Dimensions}
\label{appendix:evals}

\begin{table}[h!]
  \centering
  \small
  \begin{tabular}{m{3cm} m{12cm}}
    \toprule
    \textbf{Criteria} & \textbf{Evaluation Dimensions} \\
    \midrule
    Rationality & 
    \begin{itemize}[leftmargin=*]
      \item Whether the system accurately interprets the information provided in the user form. 
      \item Whether the flight and budget recommendations are reasonable. 
      \item Whether the budget allocation is reasonable and well-distributed. 
      \item Whether the recommendations for attractions, hotels, and restaurants are reasonable and appropriate.
      \item Whether the itinerary and schedule planning are reasonable. 
      \item Whether the number of attractions per day and the recommended visit duration are reasonable. 
      \item Whether the travel route planning is logical and efficient. 
      \item Whether special schedules are handled appropriately. 
      \item Whether adequate buffer times are included in the schedule.
    \end{itemize} \\
    \midrule
    Comprehensiveness & 
    \begin{itemize}[leftmargin=*]
      \item Whether the information provided for flights, hotels, attractions, and restaurants is detailed, rich, and up-to-date.
      \item Whether the budget planning is detailed and comprehensive. 
      \item Whether the itinerary and schedule planning are detailed.
      \item Whether the route and transportation time planning are detailed. 
      \item Whether the reasons for recommendations and planning are detailed and rich. 
      \item Whether the city introduction information and travel tips are rich. 
      \item Whether the travel planning is rich.
    \end{itemize} \\
    \midrule
    Personalization & 
    \begin{itemize}[leftmargin=*]
      \item Whether the system can perceive sufficient personal information. 
      \item Whether the flight and budget recommendations consider personal financial status. 
      \item Whether the recommendations for attractions, hotels, and restaurants consider personal preferences.
      \item Whether the intensity of itineraries is in line with personal habits. 
      
    \end{itemize} \\
    \bottomrule
  \end{tabular}
  \caption{\label{tab:human_evaluation_dimensions} Evaluation dimensions for Rationality, Comprehensiveness and Personalization used by human evaluators.}
\end{table}

\clearpage

\section{A Case Example of \method}
\label{appendix:case1}
\begin{itemize}
\item\textbf{User input form}
    
\begin{table}[h]
  \centering
  \small
  \begin{tabular}{ll}
    \toprule
    \textbf{Key} & \textbf{Value} \\
    \midrule
    user\_openai\_key & 1 \\
    user\_name & sheep \\
    departure & beijing, beijing, china \\
    destination & shanghai, shanghai, china \\
    outbound\_date & 2024-08-06 \\
    return\_date & 2024-08-09 \\
    adults\_num & 2 \\
    children\_num & 1 \\
    children\_ages & 3 \\
    restaurant\_preference & Chinese \\
    \bottomrule
  \end{tabular}
  \caption{\label{tab:json_example} JSON example for a travel scenario.}
\end{table}

\item\textbf{Overall destination description}
    
\begin{table}[h]
  \centering
  \small
  \begin{tabular}{>{\raggedright\arraybackslash}p{3cm} p{12cm}}
    \toprule
    \textbf{Key} & \textbf{Value} \\
    \midrule
    city\_name & shanghai, shanghai, china \\
    description & Shanghai, on China’s central coast, is the country's biggest city and a global financial hub. Its heart is the Bund, a famed waterfront promenade lined with colonial-era buildings. Across the Huangpu River rises the Pudong district’s futuristic skyline, including 632m Shanghai Tower and the Oriental Pearl TV Tower, with distinctive pink spheres. Sprawling Yu Garden has traditional pavilions, towers and ponds. \\
    weather & \{ "celsius": 39, "fahrenheit": 103, "conditions": "Mostly sunny" \} \\
    latitude & 31.230415999999998 \\
    longitude & 121.47370099999999 \\
    \bottomrule
  \end{tabular}
  \caption{\label{tab:city_info} City information for Shanghai.}
\end{table}

\item\textbf{Some travel tips}

\begin{tabular}{|p{14cm}|}
\hline
\begin{lstlisting}[basicstyle=\scriptsize, breaklines=true, belowskip=0pt]
"The best time to visit Shanghai is from September to early November. At this time, the weather in Shanghai is in autumn, so it is quite cool and sunny. The morning is a bit chilly, and the afternoon is still bright. Bright and warm."
"August is hot and wet all over China, even in Beijing and Xian, which receive their highest rainfall this month. Tibet is cooler than the rest of the country but still experiences rainy days. August is also one of the busiest months for the Chinese to travel with many very busy attractions."
"Stick to bottled water and only consume hot drinks during your stay. Keep your belongings close Shanghai is a safe city; however, petty crimes like pickpocketing are common. Bring toilet paper. Squat toilets aren't everywhere, but there are still some, especially in public restrooms."
"It depends on how you want to spend your time in Shanghai. Renting a car and having fun mostly indoors will be okay. Other ..."
"Journey to Inner Mongolia to stay overnight in a yurt and meander through the grasslands. Travel to Chongqing and cruise the mighty Yangtze River. Originating ..."
"To avoid the heat of summer and the cold of winter, spring (March to May) and autumn (September to November) are the best times for traveling in ..."
"It is best to avoid Shanghai in August unless you wear as little as possible and do not mind the looks by being such an obvious tourist in ..."
\end{lstlisting} \\
\hline
\end{tabular}

\clearpage

\item\textbf{Some recommended outbound flights}

\begin{tabular}{|>{\scriptsize}m{14cm}|}
\hline
\begin{lstlisting}[basicstyle=\scriptsize, breaklines=true,  belowskip=0pt]
{"id":49,"flight_number":"CZ 8879","outbound_or_return":"outbound","round_trip_price":1209,"type":"Round trip","travel_class":"Economy","total_duration":130,"departure_time":"2024-08-06 08:00","arrival_time":"2024-08-06 10:10","departure_airport":{"name":"Beijing Daxing International Airport","id":"PKX"},"arrival_airport":{"name":"Shanghai Hongqiao International Airport","id":"SHA"},"layovers":"[]","airplane":"Airbus A350","airline":"China Southern"}
\end{lstlisting} \\
\begin{lstlisting}[basicstyle=\scriptsize, breaklines=true,   aboveskip=0pt, belowskip=0pt]
{"id":50,"flight_number":"CZ 8889","outbound_or_return":"outbound","round_trip_price":1209,"type":"Round trip","travel_class":"Economy","total_duration":150,"departure_time":"2024-08-06 11:20","arrival_time":"2024-08-06 13:50","departure_airport":{"name":"Beijing Daxing International Airport","id":"PKX"},"arrival_airport":{"name":"Shanghai Pudong International Airport","id":"PVG"},"layovers":"[]","airplane":"Airbus A321","airline":"China Southern"}
\end{lstlisting} \\
\begin{lstlisting}[basicstyle=\scriptsize, breaklines=true,   aboveskip=0pt, belowskip=0pt]
{"id":51,"flight_number":"CZ 8887","outbound_or_return":"outbound","round_trip_price":1209,"type":"Round trip","travel_class":"Economy","total_duration":130,"departure_time":"2024-08-06 12:00","arrival_time":"2024-08-06 14:10","departure_airport":{"name":"Beijing Daxing International Airport","id":"PKX"},"arrival_airport":{"name":"Shanghai Hongqiao International Airport","id":"SHA"},"layovers":"[]","airplane":"Airbus A330","airline":"China Southern"}
\end{lstlisting} \\
\begin{lstlisting}[basicstyle=\scriptsize, breaklines=true,   aboveskip=0pt, belowskip=0pt]
{"id":52,"flight_number":"CZ 8885","outbound_or_return":"outbound","round_trip_price":1209,"type":"Round trip","travel_class":"Economy","total_duration":140,"departure_time":"2024-08-06 16:00","arrival_time":"2024-08-06 18:20","departure_airport":{"name":"Beijing Daxing International Airport","id":"PKX"},"arrival_airport":{"name":"Shanghai Hongqiao International Airport","id":"SHA"},"layovers":"[]","airplane":"Airbus A350","airline":"China Southern"}
\end{lstlisting} \\
\hline
\end{tabular}

\item\textbf{Some recommended return flights}

\begin{tabular}{|>{\scriptsize}m{14cm}|}
\hline
\begin{lstlisting}[basicstyle=\scriptsize, breaklines=true,  belowskip=0pt]
{"id":64,"flight_number":"CZ 8888","outbound_or_return":"return","round_trip_price":1209,"type":"Round trip","travel_class":"Economy","total_duration":135,"departure_time":"2024-08-09 15:30","arrival_time":"2024-08-09 17:45","departure_airport":{"name":"Shanghai Hongqiao International Airport","id":"SHA"},"arrival_airport":{"name":"Beijing Daxing International Airport","id":"PKX"},"layovers":"[]","airplane":"Airbus A321neo","airline":"China Southern"}
\end{lstlisting} \\
\hline
\end{tabular}

\item\textbf{Budget recommendation}

\begin{tabular}{|>{\scriptsize}m{14cm}|}
\hline
\begin{lstlisting}[basicstyle=\scriptsize, breaklines=true,  belowskip=0pt]
{"recommended_fund":"$806.50","recommendation_reason":"The estimate includes mid-range accommodation, average dining costs for 2 adults and a child, public transportation with additional taxi fares, and entry fees for popular attractions. Assumptions are based on current pricing norms for a mid-range budget in Shanghai.","round_trip_flights_price":1209}
\end{lstlisting} \\
\hline
\end{tabular}

\item\textbf{Budget allocation}

\begin{table}[h]
  \centering
  \small
  \begin{tabular}{>{\raggedright\arraybackslash}p{2.5cm} p{1cm} p{10.5cm}}
    \toprule
    \textbf{Category} & \textbf{Amount} & \textbf{Reason} \\
    \midrule
    Flights Price & \$1209 & Cost of round trip flights. \\
    Accommodation & \$300 & Selected mid-range hotels that offer comfort while keeping costs reasonable. \\
    Restaurant & \$200 & For a mix of dining experiences, including some nicer meals and budget-friendly options. \\
    Attraction & \$150 & Prioritized must-see attractions with consideration for child-friendly activities. \\
    Transportation & \$150 & Budget covers taxis and public transportation, with some room for convenience. \\
    Other Expenses & \$100 & Set aside for souvenirs and unexpected expenses. \\
    Reserve & \$100 & Reserved 10\% of the budget for emergencies and unplanned costs. \\
    \bottomrule
  \end{tabular}
  \caption{\label{tab:fund_allocation}Example of fund allocation with reasons.}
\end{table}
\clearpage

\item\textbf{Some recommended attractions}

\begin{tabular}{|>{\scriptsize}m{14cm}|}
\hline
\begin{lstlisting}[basicstyle=\scriptsize, breaklines=true,  belowskip=0pt]
{"index": 0, "image": "https://lh5.googleusercontent.com/p/AF1QipN3freQ3RdimftW-23gw96nXPQKcAZFSy1unYkm=w600-h450-k-no", "sight_name": "Yu Garden", "details": {"sight_name": "Yu Garden", "type": "['Tourist attraction', 'Attraction', 'Botanical garden']", "address": "279 Yu Yuan Lao Jie, Huangpu, Shanghai, China, 200000", "latitude": 31.2272355, "longitude": 121.492094, "per_price": "$4.16", "overall_rating": 4.5, "description": "Winding, famed classical garden. 5-acre garden built in 1577 that features Ming dynasty pavilions, ponds, rockeries & arched bridges.", "operating_hours": null, "image": "https://lh5.googleusercontent.com/p/AF1QipN3freQ3RdimftW-23gw96nXPQKcAZFSy1unYkm=w600-h450-k-no", "phone": "+86 21 6326 0830", "llm_recommend_reason": "Yu Garden is a beautiful and affordable classical garden that is suitable for all ages, offering a peaceful experience amidst the city. It's a great place for a family to explore traditional Chinese architecture and nature.", "llm_recommend_duration": 2, "llm_recommend_play_time": "[8, 12]"}, "id": 7}
\end{lstlisting} \\
\begin{lstlisting}[basicstyle=\scriptsize, breaklines=true,   aboveskip=0pt, belowskip=0pt]
{"index": 1, "image": "https://lh5.googleusercontent.com/p/AF1QipMrRvItGMkgV23nDN1fdVjdFcg_ZWPU1hoqOOp5=w600-h450-k-no", "sight_name": "People's Square", "details": {"sight_name": "People's Square", "type": "['Park']", "address": "Huang Pu Qu, China, 200003", "latitude": 31.22967, "longitude": 121.47616099999999, "per_price": "Free", "overall_rating": 4.3, "description": "Bustling square & meeting place. Urban meeting place convenient to city landmarks with a busy square & shaded garden areas.", "operating_hours": null, "image": "https://lh5.googleusercontent.com/p/AF1QipMrRvItGMkgV23nDN1fdVjdFcg_ZWPU1hoqOOp5=w600-h450-k-no", "phone": null, "llm_recommend_reason": "As a free urban park, People's Square is perfect for a family outing where you can enjoy a leisurely walk, relax, and let child play in the shaded areas. It's a central spot that's close to other attractions, making it a convenient place to visit.", "llm_recommend_duration": 1.5, "llm_recommend_play_time": "[10, 16]"}, "id": 8}
\end{lstlisting} \\
\hline
\end{tabular}

\item\textbf{Final travel plan}

\begin{tabular}{|>{\scriptsize}m{14cm}|}
\hline
\begin{lstlisting}[basicstyle=\scriptsize, breaklines=true, belowskip=0pt]
August 6, 2024
08:00 to 10:10: Outbound flight (CZ 8879 from Beijing, Beijing, China to Shanghai, Shanghai, China).
Flight Info:
- Flight Number: CZ 8879
- Airline: China Southern
- Airplane: Airbus A350
- Departure Airport: Beijing Daxing International Airport
- Arrival Airport: Shanghai Hongqiao International Airport
- Total Duration: 130 minutes
- Round Trip Price: $1209.00

10:10 to 14:00: Check-in at Dayin International Youth Hostel.
Hotel Info:
- Name: Dayin International Youth Hostel
- Address: [Full address]
- Check-in Time: 2:00 PM
- Check-out Time: 12:00 PM
- Total Price: $233
- Rating: 4.0
- Amenities: [List of amenities]

16:00 to 18:00: Have dinner at Xiao Yang Sheng Jian Huanghe Road.
Restaurant Info:
- Name: Xiao Yang Sheng Jian Huanghe Road
- Address: China, Shanghai, Huangpu, People's Square, Huanghe Road 97, 200003
- Price: $
- Rating: 4.3
- Phone: +86 21 5375 1793

18:00 to 18:10: Travel to Waibaidu Bridge.

18:10 to 18:40: Visit Waibaidu Bridge.
Sight Info:
- Name: Waibaidu Bridge
- Address: Waitan, Huang Pu Qu, China, 200002
- Rating: 4.5
- Description: Iconic all-steel bridge with LED lights. This landmark is China's first all-steel bridge & features a colorful LED lighting system.
- Recommended Duration: 0.5 hours

18:40 to 18:50: Return to Dayin International Youth Hostel.
\end{lstlisting} \\
\hline

\end{tabular}
\clearpage
\begin{tabular}{|>{\scriptsize}m{14cm}|}
\hline
\begin{lstlisting}[basicstyle=\scriptsize, breaklines=true, belowskip=0pt]
August 7, 2024
09:00 to 09:10: Travel to Yu Garden.

09:10 to 11:10: Visit Yu Garden.
Sight Info:
- Name: Yu Garden
- Address: 279 Yu Yuan Lao Jie, Huangpu, Shanghai, China, 200000
- Price: $4.16
- Rating: 4.5
- Description: Winding, famed classical garden. 5-acre garden built in 1577 that features Ming dynasty pavilions, ponds, rockeries & arched bridges.
- Recommended Duration: 2 hours

11:10 to 13:10: Have lunch at Yang's Fried Dumpling.
Restaurant Info:
- Name: Yang's Fried Dumpling
- Address: 178 Ningbo Road, Huangpu, Shanghai, China, 200002
- Price: $
- Rating: 4.4
- Phone: +86 21 6322 9763

13:10 to 13:20: Travel to Shanghai Ocean Aquarium.

13:20 to 15:20: Visit Shanghai Ocean Aquarium.
Sight Info:
- Name: Shanghai Ocean Aquarium
- Address: 1388 Lujiazui Ring Road, Lujiazui, Pudong, Shanghai, China, 200031
- Price: $22.28
- Rating: 4.2
- Description: Seals, sharks & water dragons on display. Contemporary aquarium with local & exotic marine life & long, glass-enclosed underwater walkways.
- Recommended Duration: 2 hours

15:20 to 15:30: Travel to People's Park.

15:30 to 16:30: Visit People's Park.
Sight Info:
- Name: People's Park
- Address: 231 Nanjing Rd (W), People's Square, Huangpu, Shanghai, China, 200003
- Rating: 4.4
- Description: Public green space with lawns & a pond. Set amid skyscrapers, this urban park offers landscaped lawns & wooded areas, plus a pond & cafe.
- Recommended Duration: 1 hour

16:30 to 18:30: Have dinner at China Wok.
Restaurant Info:
- Name: China Wok
- Address: 775 W Golf Rd, Des Plaines, IL 60016
- Price: $
- Rating: 4.1
- Phone: (847) 952-1688

18:30 to 18:40: Return to Dayin International Youth Hostel.
\end{lstlisting} \\
\hline
\end{tabular}
\clearpage

\begin{tabular}{|>{\scriptsize}m{14cm}|}
\hline
\begin{lstlisting}[basicstyle=\scriptsize, breaklines=true, belowskip=0pt]
August 8, 2024
09:00 to 09:10: Travel to People's Square.

09:10 to 10:10: Visit People's Square.
Sight Info:
- Name: People's Square
- Address: Huang Pu Qu, China, 200003
- Price: Free
- Rating: 4.3
- Description: Bustling square & meeting place. Urban meeting place convenient to city landmarks with a busy square & shaded garden areas.
- Recommended Duration: 1 hour

10:10 to 12:10: Have lunch at Lost Heaven Bar.
Restaurant Info:
- Name: Lost Heaven Bar
- Address: 17 Yan'an East Road, Waitan, Huangpu, Shanghai, China, 200002
- Price: $$
- Rating: 4.2
- Phone: +86 21 6330 0967

12:10 to 12:20: Travel to Former Residence of Sun Yat-Sen.

12:20 to 13:20: Visit Former Residence of Sun Yat-Sen.
Sight Info:
- Name: Former Residence of Sun Yat-Sen
- Address: 7 Xiangshan Road, Luwan District, China, 200041
- Rating: 4.3
- Description: Former home of revolutionary Sun Yat-sen. This house, where revolutionary Sun Yat-sen lived from 1918 to his death, has artifacts & a statue.
- Recommended Duration: 1 hour

13:20 to 13:30: Travel to Jade Buddha Temple.

13:30 to 14:30: Visit Jade Buddha Temple.
Sight Info:
- Name: Jade Buddha Temple
- Address: 170 Anyuan Road, Jing'an District, China, 200061
- Rating: 4.6
- Description: Monastery with many sculptures. Active Buddhist temple known for its large, ornate jade Buddha sculptures.
- Recommended Duration: 1 hour

14:30 to 14:40: Travel to The Bund.

14:40 to 16:10: Visit The Bund.
Sight Info:
- Name: The Bund
- Address: Zhongshan Rd (E-1), Waitan, Huangpu, Shanghai, China, 200002
- Price: Free
- Rating: 4.7
- Description: Famed waterside walkway with city views. Historic, 1,500-meter-long waterside promenade showcasing distinctive modern & colonial buildings.
- Recommended Duration: 1.5 hours

16:10 to 18:10: Have dinner at Jan Mei Chinese Takeout & Delivery.
Restaurant Info:
- Name: Jan Mei Chinese Takeout & Delivery
- Address: 8905 W Oakland Park Blvd, Sunrise, FL 33351
- Price: $
- Rating: 4.4
- Phone: (954) 749-0699

18:10 to 18:20: Return to Dayin International Youth Hostel.
\end{lstlisting} \\
\hline
\end{tabular}

\clearpage
\begin{tabular}{|>{\scriptsize}m{14cm}|}
\hline
\begin{lstlisting}[basicstyle=\scriptsize, breaklines=true, belowskip=0pt]
August 9, 2024
09:00 to 11:25: Hotel check-out and prepare for the return flight.
Hotel Info:
- Name: Dayin International Youth Hostel
- Address: [Full address]
- Check-in Time: 2:00 PM
- Check-out Time: 12:00 PM
- Total Price: $233
- Rating: 4.0
- Amenities: [List of amenities]

11:25 to 13:40: Return flight (CZ 8880 from Shanghai, Shanghai, China to Beijing, Beijing, China).
Flight Info:
- Flight Number: CZ 8880
- Airline: China Southern
- Airplane: Airbus A350
- Departure Airport: Shanghai Hongqiao International Airport
- Arrival Airport: Beijing Daxing International Airport
- Total Duration: 135 minutes
- Round Trip Price: $1209.00
\end{lstlisting} \\
\hline
\end{tabular}

\end{itemize}

\clearpage
\section{A Case Example of GPT-4+ agent (Baseline)}
\label{appendix:case2}

\begin{itemize}
\item\textbf{User input form}
    
\begin{table}[h]
  \centering
  \small
  \begin{tabular}{ll}
    \toprule
    \textbf{Key} & \textbf{Value} \\
    \midrule
    user\_openai\_key & 1 \\
    user\_name & sheep \\
    departure & beijing, beijing, china \\
    destination & shanghai, shanghai, china \\
    outbound\_date & 2024-08-06 \\
    return\_date & 2024-08-09 \\
    adults\_num & 2 \\
    children\_num & 1 \\
    children\_ages & 3 \\
    restaurant\_preference & Chinese \\
    \bottomrule
  \end{tabular}
  \caption{\label{tab:json_example} JSON example for a travel scenario.}
\end{table}

\item\textbf{Final travel plan}

\begin{tabular}{|>{\scriptsize}m{14cm}|}
\hline
\begin{lstlisting}[basicstyle=\scriptsize, breaklines=true, belowskip=0pt]
August 6, 2024
**Morning**
- Depart from Beijing: Take a morning high-speed train from Beijing South Railway Station to Shanghai Hongqiao Railway Station. Travel time is approximately 4.5 hours.
Activity: Enjoy scenic views of the countryside and prepare a small travel kit to keep your child entertained.
**Afternoon**
- Check into hotel: Recommended family-friendly hotels for a magical experience with child-friendly amenities.
- Lunch: Dine at Din Tai Fung in Xintiandi, known for its child-friendly environment and delicious Chinese cuisine.
**Evening**
- Explore The Bund: Take a leisurely walk along the Bund to enjoy the stunning skyline views of Shanghai.
- Dinner: Enjoy a meal at Hakkasan Shanghai for exquisite Chinese dishes in a luxurious setting suitable for families.
\end{lstlisting} \\
\hline
\end{tabular}

\begin{tabular}{|>{\scriptsize}m{14cm}|}
\hline
\begin{lstlisting}[basicstyle=\scriptsize, breaklines=true, belowskip=0pt]
August 7, 2024
**Morning**
- Shanghai Disneyland: Spend the day at Shanghai Disneyland. Pre-book your tickets and plan to visit children-friendly attractions like Fantasyland and Toy Story Land.
- Breakfast: Inside the park at Remy's Patisserie for quick bites.
**Afternoon**
- Continue at Disneyland: Explore more shows and parades suitable for your child.
- Lunch: At the Royal Banquet Hall in Disneyland, featuring character dining experiences.
**Evening**
- Return to Hotel: Rest and refresh at your hotel.
- Dinner: Explore local eateries near your hotel for authentic Shanghai cuisine.
\end{lstlisting} \\
\hline
\end{tabular}

\begin{tabular}{|>{\scriptsize}m{14cm}|}
\hline
\begin{lstlisting}[basicstyle=\scriptsize, breaklines=true, belowskip=0pt]
August 8, 2024
**Morning**
- Yu Garden: Visit the beautiful and historic Yu Garden, a peaceful place for a family stroll.
- Breakfast: Traditional Shanghai breakfast near Yu Garden with selections like xiaolongbao and congee.
**Afternoon**
- Shanghai Science and Technology Museum: Ideal for children and adults alike.
- Lunch: Casual dining at the museum's cafe, offering Chinese and Western options.
**Evening**
- Dinner: Choose from various local food stalls along Nanjing Road, offering quick and tasty options.
\end{lstlisting} \\
\hline
\end{tabular}

\begin{tabular}{|>{\scriptsize}m{14cm}|}
\hline
\begin{lstlisting}[basicstyle=\scriptsize, breaklines=true, belowskip=0pt]
August 9, 2024
**Morning**
- Pack up: Prepare for your journey back to Beijing.
**Noon**
- Travel Back to Beijing: Head to Shanghai Hongqiao Railway Station for your return trip to Beijing.
\end{lstlisting} \\
\hline
\end{tabular}
\end{itemize}

\end{appendix}

\end{document}